\documentclass[runningheads]{llncs}
\usepackage[T1]{fontenc}
\usepackage{graphicx}
\usepackage{booktabs}
\usepackage[misc]{ifsym}
\usepackage{amsmath,amssymb}
\usepackage{mwe}
\newcommand{\corr}{(\Letter)}

\begin{document}

\title{Causal Graphs Meet Thoughts: Enhancing Complex Reasoning\\
       in Graph-Augmented LLMs}
\titlerunning{Causal Graphs Meet Thoughts}

\author{
Hang Luo\inst{1}\corr \and
Jian Zhang\inst{2} \and
Chujun Li\inst{3}
}

\authorrunning{H. Luo et al.}

\institute{
Department of Computer Science, University of Warwick, Coventry CV4 7EQ, UK\\
\email{hang.luo@warwick.ac.uk}
\and
Zhejiang University, Hangzhou, Zhejiang 310058, China\\
\email{zj2001@zju.edu.cn}
\and
Department of Computer Science, Beijing University of Posts and Telecommunications, Beijing 100876, China\\
\email{lichujun@bupt.edu.cn}
}
\maketitle
\vspace{-0.2cm}
\begin{abstract}
	In knowledge-intensive reasoning tasks, especially in high-stakes domains like medicine and law, there are high demands for both the accuracy of information retrieval and the causal reasoning and explainability of the reasoning process. Large language models, due to their powerful reasoning abilities, have demonstrated remarkable potential in knowledge-intensive tasks. However, they still face challenges such as difficulty in incorporating new knowledge, generating hallucinations, and explaining their reasoning process. To address these challenges, integrating knowledge graphs with Graph Retrieval-Augmented Generation (Graph-RAG) has emerged as an effective solution. Traditional Graph-RAG methods often rely on simple graph traversal or semantic similarity, which do not capture causal relationships or align well with the model's internal reasoning steps. This paper proposes a novel pipeline that filters large knowledge graphs to emphasize cause-effect edges, aligns the retrieval process with the model's chain-of-thought (CoT), and enhances reasoning through multi-stage path improvements. 
  Experiments on medical question-answering tasks demonstrate consistent improvements in diagnostic accuracy, with up to a 10\% absolute gain across multiple large language models (LLMs), evaluated on two medical domain datasets. This approach demonstrates the value of combining causal reasoning with stepwise retrieval, leading to more interpretable and logically grounded solutions for complex queries.
\end{abstract}

\keywords{Graph retrieval-augmented generation \and Causal reasoning \and Knowledge graph \and Large language models \and Chain-of-thought.}

\section{Introduction}
\label{sec:intro}
Human civilization has always advanced by systematically gathering and applying knowledge. In contemporary settings, this process typically involves integrating diverse information from large text corpora, databases, and real-world observations. Domains such as healthcare, law, and scientific research illustrate the increasing complexity of this task: clinicians correlate patient records, guidelines, and published findings to make diagnostic decisions; legal professionals cross-reference statutes, precedents, and factual evidence to build robust arguments; and scientists synthesize large volumes of experimental data to propose novel hypotheses. The ability to \emph{retrieve}, \emph{analyze}, and \emph{reason} across multiple sources collectively known as \emph{knowledge-intensive queries} is essential for achieving precision, clarity, and innovation.
\subsection{Retrieval Challenges in Knowledge-Intensive Queries}
In Knowledge-Intensive Queries tasks, significant progress has been made using retrieval-augmented paradigms where queries are matched to semantically related documents~\cite{3,4}. 
However, many knowledge-intensive tasks demand \emph{more than} semantic overlap: 
they require understanding \emph{logical structure} and \emph{causal} pathways that connect pieces of evidence~\cite{5}. 
For instance, retrieving an article stating ``Disease~X is associated with Gene~Y'' may be thematically relevant but might not fully answer \emph{how} or \emph{why} Gene~Y leads to a specific disease manifestation. 
Hence, purely correlation-based retrieval can fall short in scenarios where an explicit explanation or directional link is needed.

Knowledge graphs, as an effective structural knowledge carrier, are often used in retrieval tasks. However, large knowledge graphs (KGs) can exacerbate retrieval complexity. As the number of entities and edges grows, often mixing correlational or loosely defined relationships-simple similarity-based traversals may retrieve extensive sets of partially relevant or conflicting facts~\cite{7new1}. Systems relying solely on dense or symbolic retrieval can thus accumulate large volumes of loosely associated data, lacking a robust mechanism for highlighting causal edges that yield a coherent inference path, frequently resulting in incomplete or contradictory answers.

Recent large language models (LLMs) have demonstrated remarkable capabilities in natural language generation and single-step knowledge retrieval~\cite{8,9}. 
Yet, these models often struggle to maintain consistent reasoning across multiple inference steps, which can lead to hallucinations or logically inconsistent conclusions~\cite{10}. 
Without a transparent mechanism to explain \emph{why} certain pieces of information connect to others, 
LLM-based pipelines cannot reliably ensure that each retrieval step aligns with an evolving chain of reasoning~\cite{11}. 
This situation makes it difficult to confirm the correctness of multi-stage answers and raises interpretability and reliability concerns.

An emerging consensus holds that \emph{causality} is a critical missing piece in knowledge-intensive tasks~\cite{12,13}. 
Correlation-based retrieval (e.g., "A relates to B") may suffice for thematically aligned lookups but fails to encode \emph{directional} or \emph{explanatory} structures. 
By contrast, causal relationships indicate \emph{how} and \emph{why} one factor influences another~\cite{15}, 
an especially crucial requirement in medical, legal, and scientific problem-solving.

Furthermore, \emph{chain-of-thought} (CoT) prompting~\cite{11} improves the interpretability of large language models 
by allowing them to articulate intermediate steps of reasoning. 
Rather than relying on the latent parameter space to form an opaque conclusion, CoT can reduce illusions of correctness 
and highlight each sub-decision, thereby improving reliability and explainability. 
When paired with a structured retrieval mechanism, CoT yields a more dynamic and context-aware process, 
since it indicates precisely \emph{which} facts the model needs at each step for a coherent solution.

\subsection{Key Limitations in Existing Graph-RAG Approaches}
\label{sec:key_limitations}

While graph-based retrieval-augmented generation (RAG) has delivered improvements
in various knowledge-intensive reasoning tasks~\cite{19new1,19new2}, the following interlocking challenges
continue to undermine its reliability and explainability---especially in complex
knowledge domains such as healthcare:

\begin{enumerate}
\item \textbf{Lack of Explicit Causal Priorities:}  
  Many existing systems treat \emph{all edges} as equally important or relevant, overshadowing the \emph{truly causal} links amid a sea of purely semantic or correlational relations. 
  These relations lack directionality or explanatory power.  
  \\
  Because the retrieval engine cannot distinguish strong cause-effect edges from weaker associative edges, 
  the final retrieval often includes tangential or trivial facts. As a result:
  \begin{itemize}
    \item \emph{Path Confusion:} The system's final reasoning path may jump between loosely connected nodes, 
    making it hard for the model to isolate the true \emph{mechanism} underlying the question.  
    \item \emph{Contradictory or Redundant Evidence:} Over-representation of non-causal edges leads to clutter, 
    forcing the model's inference steps to process conflicting or irrelevant details, increasing the chance of logical error.  
    \item \emph{Reduced Explainability:} Without focusing on cause-effect relationships, the model's \emph{explanations} 
    (or chain-of-thought) become generic or incoherent, hampering trust in the system for high-stakes domains.
  \end{itemize}

\item \textbf{Limited Alignment with Model Reasoning:}  
  In many designs, the retrieval pipeline operates \emph{independently} from the LLM's step-by-step inference~\cite{11}. 
  Even if the LLM internally follows a coherent chain-of-thought, the external retrieval system might supply large batches of data 
  unrelated to the current inference step.  
  \\
  This misalignment severely undermines the synergy between \emph{how the model thinks} 
  and \emph{which facts are fetched}:
  \begin{itemize}
    \item \emph{Disconnected or Contradictory Support:} The LLM's partial solution steps cannot consistently refine retrieval 
    because the engine is unaware of the next sub-question. The model thus tries to integrate data that may not fit 
    the immediate line of reasoning.  
    \item \emph{Erratic or Inconsistent Conclusions:} Since the intermediate logic does not match the retrieved evidence, 
    the final answer can appear correct in some steps but then conflict with half of the retrieved context, 
    causing contradictory or ambiguous reasoning.  
    \item \emph{Lost Opportunity for Iterative Querying:} The model's chain-of-thought typically divides the query 
    into sub-questions; a disconnected retrieval pipeline cannot capitalize on these sub-questions to progressively 
    narrow down relevant nodes and edges.
  \end{itemize}

\item \textbf{Challenges of KG Overgrowth:}  
  State-of-the-art KGs often contain \emph{millions} of edges, with multiple alternative routes linking the same entities~\cite{1new}. 
  While this breadth can be advantageous for coverage, it complicates the identification of a succinct subgraph 
  for any particular question.  
  \\
  The sheer scale and connectivity of the KG can lead to:
  \begin{itemize}
    \item \emph{Excessive Graph Noise:} With so many edges often correlational or tangential, the system must filter out 
    large quantities of low-utility data. If the filtering is inadequate or too lenient, the correct cause-effect path 
    is buried under irrelevant expansions.  
    \item \emph{Diluted Reasoning Path:} Even if a correct path \emph{exists}, the model might merge it with numerous spurious edges, 
    forming an oversized retrieval set. This \emph{dilution} confuses the chain-of-thought logic, potentially generating 
    contradictory or incomplete answers.  
    \item \emph{Scalability Pressures:} Searching for multi-hop paths in a densely interconnected graph becomes computationally expensive, 
    raising practical concerns for real-time or large-scale QA applications.
  \end{itemize}
\end{enumerate}

\paragraph{Significance and Motivation:}  
Given these pitfalls, advanced methods must emphasize:
\begin{enumerate}
\item \textbf{Causal-first retrieval}, ensuring that strong cause-effect links are \emph{prioritized} over weaker or purely associative edges.  
\item \textbf{Reasoning-driven alignment}, allowing the LLM's chain-of-thought to \emph{steer} retrieval queries at each intermediate inference step, 
thus bridging the gap between how the model reasons and how the system fetches evidence.  
\item \textbf{Noise mitigation and subgraph pruning}, focusing on removing unhelpful or misleading expansions within large KGs, 
so that the final evidence set remains coherent and comprehensible.
\end{enumerate}
Addressing these aspects is crucial for delivering reliable knowledge-intensive reasoning tasks especially in domains like medicine, 
where even small misalignments or spurious edges can produce dangerously incorrect recommendations. 

\subsection{Contributions and Paper Organization.}
In this framework, two principles hold equal importance: (1) \emph{prioritizing causal edges} within a large knowledge graph, 
and (2) \emph{aligning retrieval with the LLM's chain-of-thought} to dynamically guide each reasoning step. 
Concretely, a hierarchical retrieval paradigm is proposed that looks for \emph{strong cause-effect links first} 
and resorts to broader correlational or semantic edges \emph{only if} purely causal paths do not suffice. 
Termed \emph{Causal-First Graph-RAG}, this approach ensures that potential cause-effect chains are explored early, 
while still preserving a fallback route for coverage.

The framework of this research is as follows:
\begin{enumerate}
\item Develops a hierarchical retrieval scheme prioritizing cause-effect edges, thereby mitigating the noise of large KGs 
and bridging the gap between correlation-based retrieval and truly \emph{cause-oriented} reasoning;
\item Integrates chain-of-thought prompting to precisely align each reasoning step with graph lookups, 
ensuring retrieval remains in sync with the evolving inference process;
\item Applies a causal inference-driven filtering stage to remove contradictory or tangential sub-paths, 
retaining only the subgraph consistent with the model's chain-of-thought.
\end{enumerate}
Taken together, these contributions advance both \emph{retrieval quality} (through causal prioritization) 
and \emph{reasoning consistency} (through CoT guidance), culminating in a scalable pipeline for 
complex, knowledge-intensive reasoning tasks in medical and similarly knowledge-intensive domains.

\section{Related Works}
\label{sec:related_work}

\subsection{Retrieval-Augmented Generation and Graph-Based Knowledge Discovery}
Retrieval-Augmented Generation (RAG)~\cite{3} has become a central paradigm for embedding external knowledge into generative models, improving performance on knowledge-intensive tasks. Traditional text-based retrieval, such as dense passage retrieval (DPR)~\cite{4}, is effective for single-fact lookups but struggles with queries requiring complex knowledge integration or logical coherence. To more systematically encode relational information, a number of studies advocate 
\emph{graph-based retrieval} methods, wherein knowledge graphs (KGs) store explicit connections among entities. 
Graph-RAG systems traverse these links to assemble more directed and interpretable evidence sets. 
This strategy supports entity disambiguation, relational reasoning, and targeted knowledge selection, 
providing a structured alternative to raw text passages. 
However, most present-day graph retrieval systems rely heavily on correlation-based edges---such as 
"\emph{associated with}" or "\emph{related to}"---that do not necessarily capture the \emph{directionality} 
or \emph{causal significance} required by more demanding queries. 
As a result, systems can retrieve relevant but non-explanatory connections, 
yielding incomplete or logically inconsistent knowledge sets in scenarios that demand deeper causal or sequential reasoning.

\subsection{Causality in Knowledge-Intensive Natural Language Processing}
Causality has long been recognized as a fundamental cornerstone of advanced knowledge discovery 
in fields spanning cognitive science, epidemiology, and machine learning~\cite{14}. 
Within natural language processing (NLP), numerous lines of inquiry---from causal event extraction 
and counterfactual generation to specialized question answering---aim to move beyond simple correlations 
toward \emph{why}-oriented explanations~\cite{12,13}. 
Nevertheless, integrating explicit causal inference into retrieval-centric architectures remains nontrivial. 
Many frameworks stop at correlation-based links or unstructured text retrieval, 
lacking specialized graph structures or annotated metadata to differentiate \emph{cause-and-effect} 
from adjacency or co-occurrence~\cite{15}. 
Hence, while semantically relevant facts can be extracted, the logical consistency or directional flow 
crucial in knowledge-intensive queries is not guaranteed.

\subsection{Chain-of-Thought Prompts and Reasoning Transparency}
Chain-of-thought (CoT) prompting~\cite{11} has recently garnered attention as a method to expose a language model's intermediate reasoning path. 
By prompting the model to sequentially articulate partial conclusions rather than jumping to a final answer, 
CoT can (i) provide interpretable glimpses into the model's decision-making process and (ii) sometimes improve performance 
on tasks requiring compositional or multi-stage thinking. 
However, the widespread usage of CoT remains largely \emph{decoupled} from retrieval pipelines. 
Typically, the LLM generates an internal chain-of-thought, but the system does not subdivide retrieval according to each CoT step. 
Consequently, any fine-grained reasoning states are not matched with equally fine-grained retrieval steps, 
creating potential mismatches.

\subsection{Challenges in Aligning Causality, Graph Retrieval, and CoT Reasoning}
Various research efforts seek to enrich retrieval-augmented generation by weaving in explicit inference or causal structures~\cite{11}. 
Yet persistent difficulties remain:
\begin{enumerate}
\item \textbf{Causal Under-specification.}
  Many graph-based retrieval modules store correlations or broad associations but do not designate strong \emph{cause-and-effect} edges. 
  As a result, the pipeline fails to differentiate essential causal links from incidental relationships.
\item \textbf{Misaligned Retrieval vs.\ Reasoning Steps.}
  Chain-of-thought prompts typically divide the reasoning into multiple short sub-steps, 
  but standard Graph-RAG retrieval is organized as a single module that either returns a large chunk of data 
  or lumps all evidence together. 
  This misalignment can cause the LLM to request step-by-step clarifications, 
  yet the retrieval engine has not subdivided the knowledge base lookups accordingly, 
  thus returning content irrelevant to the \emph{current} sub-question. 
\item \textbf{Scalability Issues with Large Knowledge Graphs.}
  In real-world applications, knowledge graphs may contain millions of edges with varying specificity or utility. 
  A naive graph traversal often yields excessively large or tangentially related subgraphs, 
  eclipsing the critical edges. This surplus can confuse both causal reasoning and chain-of-thought, 
  resulting in contradictory or incomplete final answers.
\end{enumerate}

\noindent
The present study introduces a unified framework that elevates causal edges within a large-scale graph environment 
while synchronizing retrieval with the LLM's chain-of-thought states. 
By assigning higher priority to cause-effect relationships in the subgraph search, 
the pipeline avoids overloading the model with loosely associated edges. 
Furthermore, each chain-of-thought step prompts a targeted query, ensuring fine-grained alignment 
between what the LLM considers at a given moment and which graph edges or nodes are returned. 
This synergy preserves the interpretability gains of CoT while injecting causal rigor into the retrieval process, 
thereby addressing correlation-based noise and subgraph sprawl. 
\section{Methodology}
\label{sec:method}

\begin{figure}[h]
    \centering
    \includegraphics[width=\textwidth]{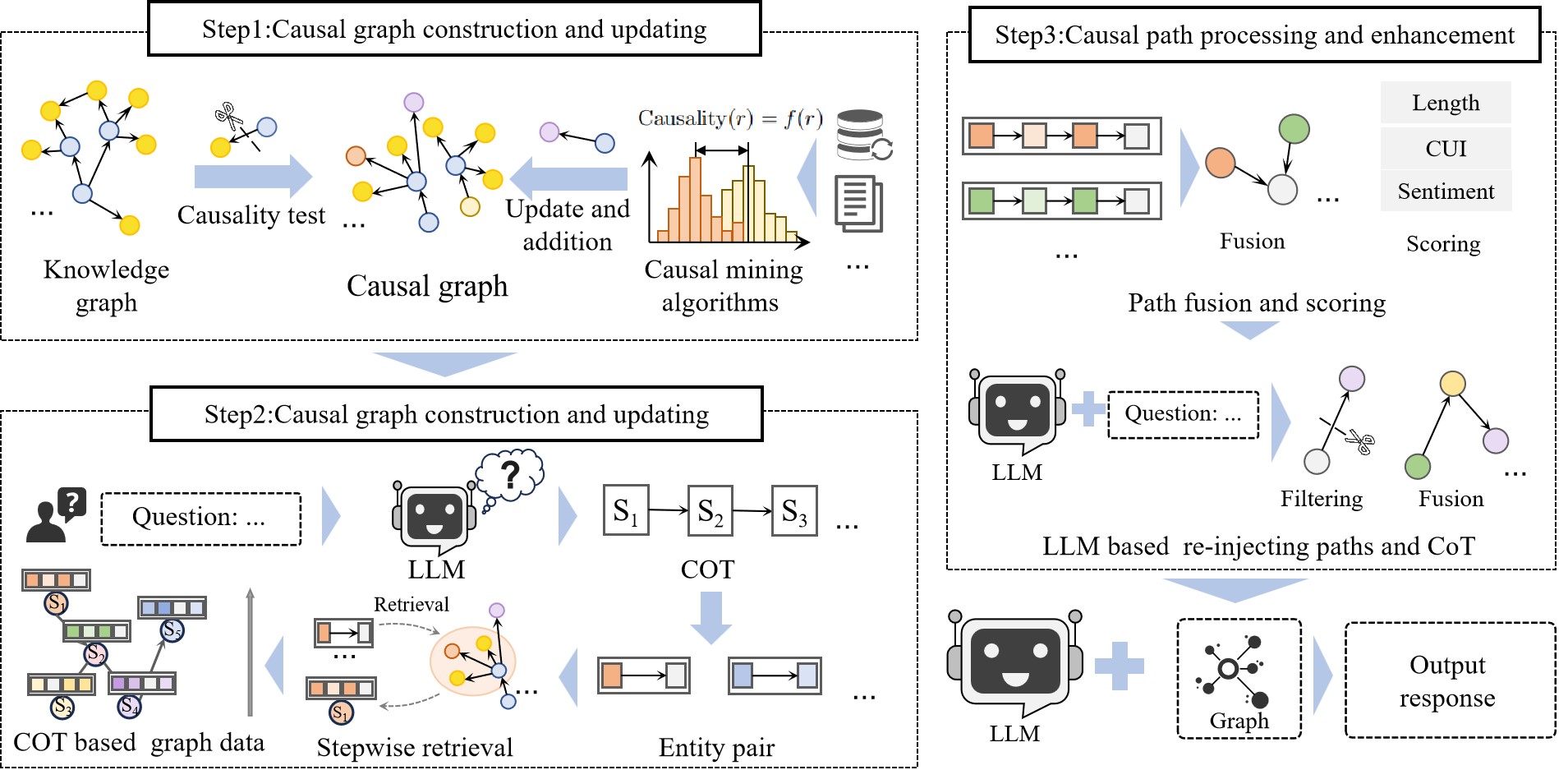}
    \caption{Overall framework in three stages:
    (1) building/updating the causal graph,
    (2) chain-of-thought-driven retrieval,
    (3) multi-stage path processing.}
    \label{fig:overall}
\end{figure}

This section describes the pipeline for 
(1) constructing a \emph{causal subgraph} from a large knowledge graph (KG), 
(2) generating a \emph{chain-of-thought} (CoT) that segments reasoning steps using "$\rightarrow$" markers, 
and (3) performing multi-stage retrieval and path enhancement with fallback strategies, including a final phase that re-injects the merged paths and original CoT into the language model. Figure~\ref{fig:overall} provides an overview.
\vspace{-0.4cm}
\vspace{-0.6cm}
\subsection{Overall Framework}
The method is divided into three main components:

\begin{enumerate}
\item \textbf{Causal graph construction and updating}: A large KG is scanned, and only edges carrying cause-effect significance are retained, assigning numeric strengths. Edges deemed purely correlational are excluded if their relevance falls below a threshold.

\item \textbf{Chain-of-thought-driven causal path retrieval}: A CoT is acquired from the LLM, splitting each segment by "$\rightarrow$," and attempting to find paths in the causal subgraph that connect the relevant entities. If no match is found, the method reverts to the full KG.

\item \textbf{Causal path processing and multi-stage enhancement}: After retrieval, structurally similar paths are merged, refined scores (based on entity/semantic overlaps and path length) are computed, and the top-ranked subset is presented as final evidence. Both the final path set and the original CoT are then fed back to the LLM for a consistency check and "enhanced" output.
\end{enumerate}

Below, each stage is detailed, referencing code snippets and the formulae used for scoring and filtering.

\subsection{Causal Graph Construction and Updating}
\label{sec:causal_construction}

\subsubsection{Filtering the Knowledge Graph for Causality.}
Let the original knowledge graph be $\mathcal{G} = (\mathcal{V}, \mathcal{E})$. Each edge $(u, r, v)\in \mathcal{E}$ has a relation label $r$ that could be \texttt{CAUSE}, \texttt{MANIFESTATION\_OF}, \texttt{ASSOCIATED\_WITH}, etc. A \emph{causality function} is defined:
\begin{equation}
  \mathrm{Causality}(r) = f(r),
\end{equation}
which assigns stronger numeric weights to relations that directly indicate cause-effect relationships (e.g., \texttt{CAUSE}). Edges whose weight $f(r)$ is below a threshold $\theta$ are discarded. Formally, the \emph{causal subgraph} $\mathcal{G}_C$ is:
\begin{equation}
\label{eq:causal_subgraph}
\mathcal{G}_C 
\;=\;
\Bigl\{\,
(u, r, v) \,\Bigm|\,
(u, r, v)\in \mathcal{E},\;\mathrm{Causality}(r)\,\ge \theta
\Bigr\}.
\end{equation}
This process excludes non-causal or weakly causal edges, leaving only those with explicit cause-effect relevance.

\subsubsection{Continuous Updates via Causal Mining.}
If new data suggests revised causal evidence for any edge, a causal mining algorithm (e.g., PC~\cite{PC}) is adopted to update edge weights. Let $S_{\text{new}}(u,v)$ be the new strength for $(u, r, v)$. If $S_{\text{new}}(u,v)\ge \theta$, that edge is added or revised in $\mathcal{G}_C$:
\begin{equation}
\label{eq:causal_update}
\mathcal{G}_C 
\;\leftarrow\;
\mathcal{G}_C \;\cup\;\bigl\{\,(u, r, v)\,\bigm|\,
S_{\text{new}}(u,v)\,\ge\,\theta \bigr\}.
\end{equation}
Thus, the causal graph evolves as additional insights emerge.

\subsection{Chain-of-Thought-Driven Causal Path Retrieval}
\label{sec:cot_retrieval}
\begin{figure}[t]
    \centering
    \includegraphics[width=0.95\textwidth]{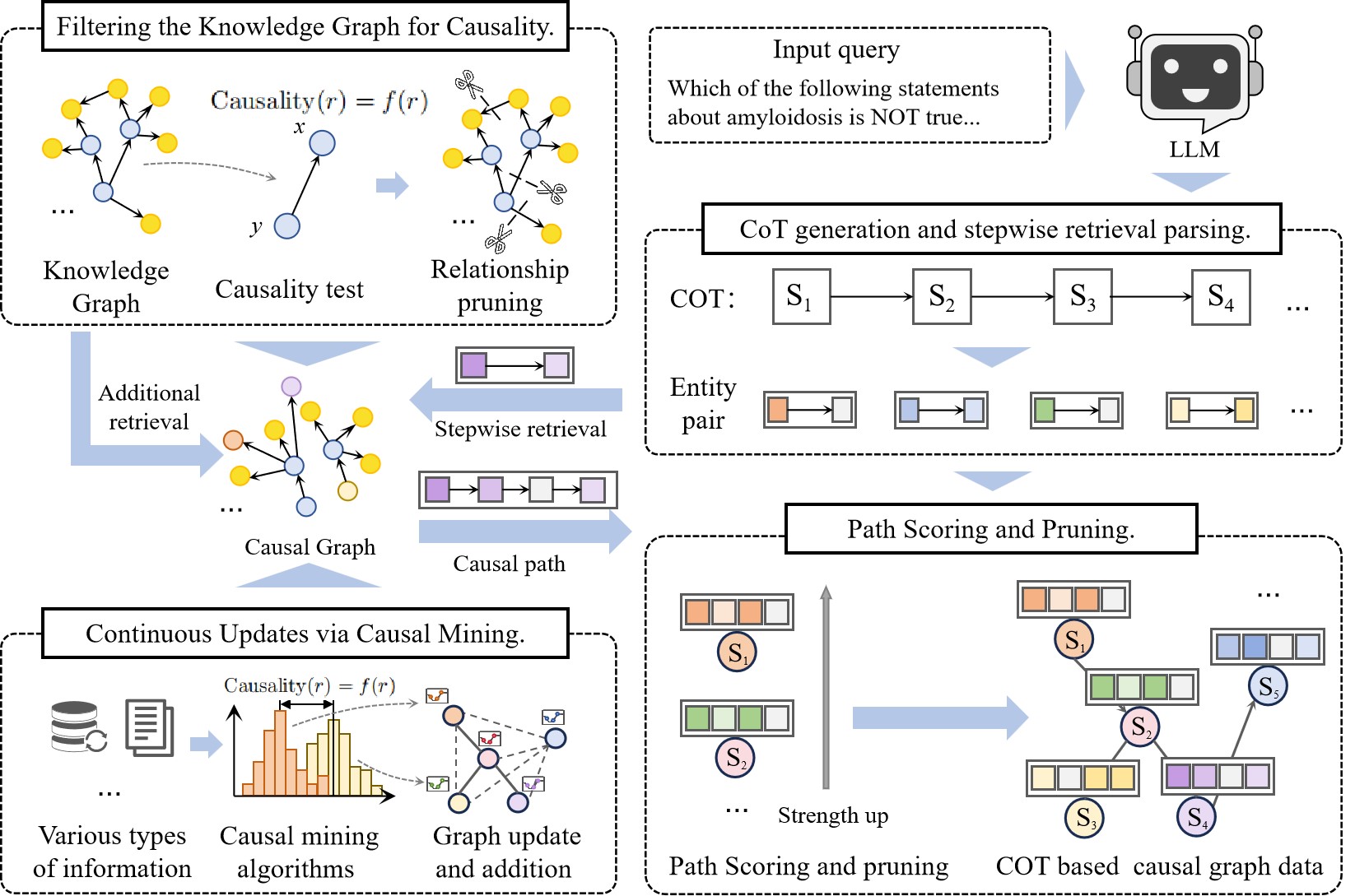}
    \caption{An example of chain-of-thought-driven retrieval. 
    Each CoT segment triggers a partial query to the causal subgraph, 
    producing candidate paths that are later fused.}
    \label{fig:cot_retrieval}
\end{figure}

This section outlines a method for causal path retrieval driven by the chain-of-thought process, which facilitates more coherent and targeted reasoning, as shown in \textbf{Figure~\ref{fig:cot_retrieval}}.

\subsubsection{CoT Generation and Parsing.}
For each query, the LLM is prompted to produce a chain-of-thought (CoT) in short segments, each separated by "$\rightarrow$" and ending with a numeric confidence in $[0,100]$. The chain is then split on these separators (omitting the final confidence) into segments $\{s_1, s_2,\dots, s_T\}$, each $s_i$ being a concise statement or state.

\subsubsection{Stepwise Retrieval per CoT Segment.}
A domain-specific entity recognizer is run on each segment $s_i$, yielding entity sets $\mathcal{E}(s_i)$. For each consecutive pair $(s_i, s_{i+1})$ in the CoT, the method attempts to connect the recognized entities of $s_i$ to those of $s_{i+1}$ by querying the \emph{causal subgraph} $\mathcal{G}_C$. If no suitable path is found, the fallback is the original KG $\mathcal{G}$. These segment-level queries produce a candidate path pool $\mathcal{P}_{\mathrm{combined}}$.

\subsubsection{Path Scoring and Pruning.}
Once $\mathcal{P}_{\mathrm{combined}}$ is assembled, a \(\mathrm{PathScore}\) is assigned to each path $p=\{e_1,\dots,e_L\}$. Let \(\mathrm{strength}(e_i)\) denote the numeric weight (causal strength if in $\mathcal{G}_C$) of edge $e_i$. The path score is computed as:
\begin{equation}
\label{eq:path_score}
\mathrm{PathScore}(p)
\;=\;
\frac{\sum_{i=1}^{L}\mathrm{strength}(e_i)}{L}.
\end{equation}
Up to $k$ top-scoring paths per segment are then selected:
\begin{equation}
\label{eq:topk_path}
\mathrm{SelectedPaths}_{\mathrm{segment}}
\;=\;
\mathrm{Top}\text{-}k
\Bigl(
\bigl\{
\mathrm{PathScore}(p)
\bigr\}_{p\,\in\,\mathcal{P}_{\text{combined}}}
\Bigr).
\end{equation}
Paths containing loops (e.g., $n_i=n_j$ for $i\neq j$) or drastically exceeding a shortest-distance threshold are pruned, ensuring that $\mathrm{SelectedPaths}_{\mathrm{segment}}$ remain concise.

\subsection{Causal Path Processing and Multi-Stage Enhancement}
\label{sec:enhancement}
\begin{figure}[h]
    \centering
    \includegraphics[width=1\textwidth]{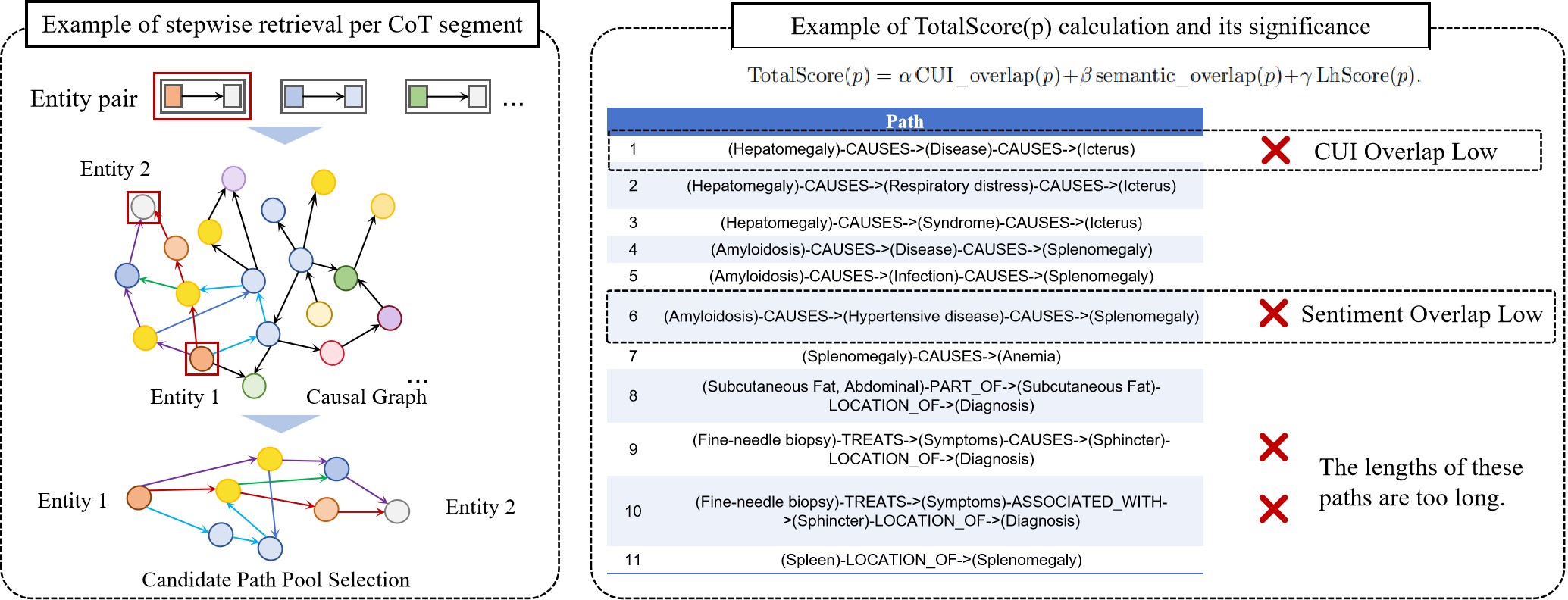}
    \caption{Path scoring and pruning. 
    Each path is evaluated based on CUI overlap, semantic overlap, and length heuristics.}
    \label{fig:scoring_example}
\end{figure}

After obtaining candidate paths from each CoT segment, 
we compute a combined score to prune irrelevant or overly long paths.
Specifically, we define $\mathrm{TotalScore}(p)$ as in Equation~(\ref{eq:total_score_with_greek}).
\textbf{Figure~\ref{fig:scoring_example}} shows a sample scenario of how 
scoring and pruning operate on three candidate paths.

\vspace{-0.6cm}
\subsubsection{First-Stage Enhancement: Path Fusion and Preliminary Scoring}
All segment-level paths (including fallback) are gathered into a global set $\mathcal{P}_{\mathrm{all}}$:
\begin{equation}
\label{eq:union_paths}
\mathcal{P}_{\mathrm{all}}
=
\bigcup_{\text{segment}}
\,\mathrm{SelectedPaths}_{\mathrm{segment}}.
\end{equation}
If multiple paths share the same start node, end node, and set of intermediate nodes, they are merged. Next, a $\mathrm{TotalScore}(p)$ is assigned to each merged path, reflecting relevance to the query. Specifically:

\paragraph{(1) CUI Overlap.}
Let $\mathcal{CUI}_{\mathrm{query}}$ be the set of CUIs from the question/answers, and $\mathcal{CUI}_{\mathrm{path}}$ that from path $p$.  
\begin{equation}
\label{eq:cui_overlap}
\mathrm{CUI\_overlap}(p)
=
\frac{
\bigl|
\mathcal{CUI}_{\mathrm{query}}
\;\cap\;
\mathcal{CUI}_{\mathrm{path}}
\bigr|
}{
\bigl|\mathcal{CUI}_{\mathrm{query}}\bigr|
}.
\end{equation}

\paragraph{(2) Semantic-Type Overlap.}
Similarly, $\mathrm{semantic\_overlap}(p)$ captures the fraction of domain-specific entity types matched between the path and query context.

\paragraph{(3) Path Length Heuristic.}
A simple length-based function is defined:
\begin{equation}
\label{eq:lhscore}
\mathrm{LhScore}(p)
=
\frac{1}{1 + \mathrm{path\_length}(p)}.
\end{equation}
Given weights $\alpha,\beta,\gamma$ with $\alpha+\beta+\gamma=1$, the method combines these into:

\begin{equation}
\label{eq:total_score_with_greek}
\mathrm{TotalScore}(p)
=
\alpha\,\mathrm{CUI\_overlap}(p)
+
\beta\,\mathrm{semantic\_overlap}(p)
+
\gamma\,\mathrm{LhScore}(p).
\end{equation}
Paths are sorted by $\mathrm{TotalScore}(p)$, and the top $\mathrm{keep\_ratio}\times|\mathcal{P}_{\mathrm{all}}|$ fraction is kept:
\begin{equation}
\label{eq:final_kept}
\mathrm{SelectedPaths}_{\mathrm{final}}
=
\mathrm{Top}
\Bigl(
\mathrm{keep\_ratio}\,\times
|\mathcal{P}_{\mathrm{all}}|
\Bigr).
\end{equation}
This step yields a semantically aligned, concise subset.
\subsubsection{Second-Stage Enhancement: Re-Injecting Paths and CoT}
After selecting the best paths, they, together with the \emph{original CoT}, are re-injected into the LLM for a final consistency check. During this phase, the model can cross-verify domain knowledge, remove spurious or contradictory branches, and produce a succinct "enhanced" summary to guide the final answer. This second-stage approach typically clarifies ambiguous steps or fuses partial sub-graphs into a coherent explanation. The final answer thus leverages both the curated paths and the chain-of-thought, improving coherence and interpretability.

\paragraph{Implementation Notes}
Each step---causal graph extraction, CoT-based retrieval, merging/scoring, and second-stage enhancement---is implemented as a separate module. Numeric \texttt{strength} values, loop checks, and shortest-path logs ensure only relevant, acyclic paths remain, with \emph{causality} prioritized in both retrieval and scoring. This multi-stage pipeline, culminating in a final re-check, harnesses the synergy between structured causal knowledge and the model's chain-of-thought for knowledge-intensive question answering.

\section{Experiments}
\label{sec:experiments}
This section evaluates the causal-first Graph-RAG framework on two medical multiple-choice QA datasets derived from MedMCQA\cite{data1} and MedQA\cite{data2}. Both datasets contain clinically oriented questions across multiple medical subtopics. Only items mapping into the causal subgraph are retained for testing; each system is compared using Precision, Recall, and F1 Score as the evaluation metrics.

\subsection{Experimental Setup}
\paragraph{Knowledge Graph.}
All retrievals occur against a SemMedDB-based knowledge graph (KG)\footnote{\url{https://lhncbc.nlm.nih.gov/ii/semmeddb/}}, containing medically oriented entities. The \emph{causal} subgraph is constructed as described in Section~\ref{sec:causal_construction}, discarding edges whose cause-effect weight is below threshold \(\theta\). Some queries that fail to match any node in this subgraph are excluded.

\paragraph{Models.}
Three large language models (LLMs) are tested:
\begin{itemize}
\item \textbf{GPT-4o}: An instruction-tuned model, consistently the top performer among those tested,
\item \textbf{GPT-4}: A similarly advanced LLM but slightly weaker in domain tasks,
\item \textbf{GPT-4o-mini} (4o-mini): A smaller-capacity variant that lags behind the others.
\end{itemize}

\subsection{Experimental Results}
\label{subsec:results}
\subsubsection{Experiment 1: Approached Method vs Direct Model Responses.}
\label{subsec:exp1}

As shown in Table 1, the LLMs leveraging the proposed method, which includes cause-effect filtering, CoT-aligned retrieval, and multi-stage path refinement, outperform direct single-step model responses across all three---Precision, Recall, and F1 Score. The results highlight that the proposed approach, particularly when applied to CGMT(GPT-4o mini), provides significant improvements over direct GPT-4o mini (with precision up to 10\% absolute), showcasing the effectiveness of this method in enhancing model performance for knowledge-intensive tasks.

\subsubsection{Experiment 2: Approached Method vs Traditional Graph-RAG.}
\label{subsec:exp2}

In the second experiment, this research compare the proposed CGMT method with traditional Graph-RAG models. As shown in Table 1, the CGMT models, particularly CGMT(GPT-4o), consistently outperform the traditional Graph-RAG models across both datasets. For example, CGMT(GPT-4o) achieves a precision of 92.90 on MedMCQA and 88.36 on MedQA, significantly higher than the corresponding values for Graph-RAG(GPT-4o) (86.89 and 83.07). This demonstrates that the proposed CGMT method, with its integrated cause-effect filtering and CoT-aligned retrieval, offers substantial improvements over traditional Graph-RAG models, especially in handling complex knowledge-intensive tasks.

\subsubsection{Experiment 3: Ablation Study}
\label{subsec:exp3}

This ablation study evaluates the individual contributions of various components of the proposed CGMT method:

\begin{itemize}
	\item \textbf{KG-only}: In this configuration, only the knowledge graph is used to search for relevant information in a correlation-based manner, without using Chain-of-Thought (CoT) prompts or causal prioritization.
	\item \textbf{Remove LLM Enhanced}: This ablation removes the final LLM-based enhancement, meaning the raw paths are merged without applying the second-stage CoT synergy.
	\item \textbf{Remove Enhancer}: This variant eliminates the entire path-enhancement step, relying solely on the retrieval process without any enhancement from the final LLM consistency pass.
\end{itemize}
\begin{table}[h]
	\centering
	\caption{Evaluation results for LLMs using the proposed method compared with traditional Graph-rag and direct responses across two datasets.}
	\label{tab:table_1}
	\resizebox{1\textwidth}{!}{  %
		\begin{tabular}{lccc|ccc}
			\toprule
			\textbf{Model} & \multicolumn{3}{c|}{\textbf{MedMCQA}} & \multicolumn{3}{c}{\textbf{MedQA}} \\
			\cmidrule(lr){2-4} \cmidrule(lr){5-7}
			& Precision & Recall &  F1 Score & Precision & Recall & F1 Score \\
			\midrule
			\textbf{CGMT(GPT-4o)}     & \textbf{92.90} &  \textbf{93.33}      &  \textbf{0.95}              & \textbf{88.36}          &  \textbf{74.03}      &   \textbf{0.84}              \\
			\textbf{CGMT(GPT-4)}      & 87.98 &  88.57      &   0.92              &  74.07         & 55.21       & 0.68                \\
			\textbf{CGMT(GPT-4o mini)}  & 82.51 &  82.67      &   0.89              &   72.49        & 53.68       &   0.66              \\
			\midrule
			\textbf{Graph-RAG(GPT-4o)}     & 86.89 &  86.81      &  0.91      & 83.07       & 70.37       & 0.81       \\
			\textbf{Graph-RAG(GPT-4)}      & 82.51 &  83.56      & 0.88       &  72.49      & 53.47       &  0.68      \\
			\textbf{Graph-RAG(GPT-4o mini)}  & 75.41 &  78.91      &  0.84      & 62.96       & 44.86       & 0.58       \\
			\midrule
			\textbf{GPT-4o}           & 85.52 &  86.30      &   0.90              & 81.48          & 65.52       & 0.78                \\
			\textbf{GPT-4}            & 84.15 &  84.83      &  0.89               & 69.84          & 51.00       & 0.64                \\
			\textbf{GPT-4o mini}        & 72.13 &   76.51     &  0.82               & 60.85          & 43.36       &  0.57               \\
			\bottomrule
		\end{tabular}
	}
\end{table}
\vspace{-0.5cm}

\begin{table}[h]
	\centering
	\caption{Ablation Study for CGMT Method with Various Enhancements Across Two Datasets.}
	\label{tab:ablation_study}
	\resizebox{1\textwidth}{!}{  %
		\begin{tabular}{lccc|ccc}
			\toprule
			\textbf{Model} & \multicolumn{3}{c|}{\textbf{Precision of MedMCQA(\%)}} & \multicolumn{3}{c}{\textbf{Precision of MedQA(\%)}} \\
			\cmidrule(lr){2-4} \cmidrule(lr){5-7}
			& GPT-4o & GPT-4 & GPT-4o mini & GPT-4o & GPT-4 & GPT-4o mini \\
			\midrule
			\textbf{CGMT}              & \textbf{92.90} & \textbf{87.98} & \textbf{83.06}  &  \textbf{88.36} &  \textbf{74.07}       &  \textbf{72.49}       \\
			\textbf{KG-only (Knowledge Graph)}   & 91.80 & 85.25 & 79.78 & 87.30       & 70.37       & 70.90       \\
			\textbf{Remove LLM Enhanced}       & 90.16 & 86.89 & 81.42 & 85.18       & 68.76       &  68.79      \\
			\textbf{Remove Enhancer}           & 90.71 & 83.61 & 78.14 & 86.24       & 71.96       &  69.84      \\
			\bottomrule
		\end{tabular}
	}
\end{table}
\vspace{-0.25cm}

As shown in Table \ref{tab:ablation_study}, the results provide insights into the contributions of each component. The full pipeline (CGMT) significantly outperforms the other variants, demonstrating the importance of each enhancement in achieving higher precision across both datasets.
\vspace{-0.2cm}

\subsection{Chain-of-Thought Cross-Model Evaluation}
\label{subsec:cot_quality}
Cross-model scenarios were tested, where one model's Chain-of-Thought (CoT) is used by another. These tests involve three stages: the first stage is CoT generation, where one model generates the CoT; the second stage is enhancement, where the CoT is refined and enhanced by a different model; and the third stage is inference, where the final answer is generated based on the enhanced CoT. Table~\ref{tab:cot_cross} details the results of three such trials across these stages. Baseline references come from the ``CGMT'' row in Table~\ref{tab:table_1}.
\subsubsection{Evaluation.} The results from the cross-model experiments provide valuable insights into the interactions between models of varying capacities. Specifically, the following observations were made:

\begin{itemize}
	\item \textbf{High-Capacity Model with Weak CoT:} When GPT-4o relies on the CoT generated by GPT-4o-mini, its accuracy drops from 92.90\% to 87.98\%. This result underscores the importance of a stronger LLM generating its own Chain-of-Thought, as the more complete reasoning chain produced by the high-capacity model leads to better performance.
	
	\item \textbf{Weak Model with Strong CoT:} In contrast, when GPT-4o-mini uses GPT-4o's CoT, its accuracy improves from 83.06\% to 85.79\%. This suggests that a more coherent and well-structured CoT can help compensate for the limited reasoning capacity of a smaller model.
	
	\item \textbf{Mixed CoT and Enhancements:} In the scenario where both CoT generation and path enhancements are carried out by the smaller model, with final inference performed by GPT-4o, the resulting performance (86.34\%) remains intermediate. This highlights the crucial role of both the initial CoT generation and the path enhancement process in determining the final outcome. Even with the final inference step handled by the higher-capacity model, the quality of the preceding stages significantly constrains overall performance.
\end{itemize}

\begin{table}[h]
	\centering
	\caption{Cross-Model CoT Usage with Three Stages (COT, Enhancement, Inference).}
	\label{tab:cot_cross}
	\resizebox{0.85\textwidth}{!}{
		\begin{tabular}{lccc}
			\toprule
			\textbf{COT} & \textbf{Enhancement} & \textbf{Inference} & \textbf{Precision of MedMCQA(\%)} \\
			\midrule
			\textbf{GPT-4o mini} & \textbf{GPT-4o}      & \textbf{GPT-4o}      & \textbf{87.98} \\
			\textbf{GPT-4o}      & \textbf{GPT-4o mini} & \textbf{GPT-4o mini} & \textbf{85.79} \\
			\textbf{GPT-4o mini} & \textbf{GPT-4o mini} & \textbf{GPT-4o}      & \textbf{86.34} \\
			\bottomrule
		\end{tabular}
	}
\end{table}
\vspace{-0.6cm}
\section{Discussion}
\label{sec:discussion}
\vspace{-0.2cm}
Although the experiments indicate that a causal-first Graph-RAG framework, aligned with chain-of-thought (CoT) prompting, can notably enhance knowledge-intensive reasoning tasks, certain limitations and open questions remain:

\paragraph{Uncertainty in Chain-of-Thought Generation.}
CoT outlines can vary under identical prompts, leading to contradictory or incomplete intermediate states if the LLM's sampling changes. This instability can produce retrieval successes or failures for the same question across multiple runs. Rarely, distinct final answers may also arise, suggesting that while CoT yields overall gains, it adds stochastic elements that must be managed.

\paragraph{Incomplete or Low-Entropy Knowledge Graphs.}
Despite leveraging a large SemMedDB-based subgraph, certain clinically relevant edges may be missing, forcing fallback retrieval from correlation-based links. The wide connectivity and shallow semantics (e.g., "related to") can also cause clutter. More robust causal-mining and curated domain expansions could mitigate these drawbacks.

\paragraph{Future Outlook.}
The results validate the advantages of aligning causal-first retrieval with CoT. Further gains may be realized by stabilizing CoT prompts (via self-consistency sampling or advanced prompt engineering) and refining the knowledge graph itself (via domain-specific expansions).

\section{Conclusion}
\label{sec:conclusion}

This work has presented a comprehensive framework integrating \emph{causal-first} knowledge graph filtration with \emph{chain-of-thought} (CoT) guided retrieval, culminating in a multi-stage path enhancement strategy. Departing from correlation-based Graph-RAG, the method prioritizes cause-effect edges in a large, densely connected medical KG and aligns retrieval steps with the model's CoT prompts by segment. Empirical evaluation on medical QA tasks (MedMCQA and MedQA) shows consistent accuracy gains across three LLMs (GPT-4o, GPT-4, GPT-4o-mini). Ablations verify that each component, from causal graph construction to second-stage LLM refinement, contributes to final performance. Cross-model CoT experiments highlight that \emph{reasoning trace quality} strongly determines downstream success. The key takeaways from this work are as follows:
\begin{itemize}
\item \textbf{Causality as a First-Class Citizen:} By explicitly ranking cause-effect edges above generic or associative links, the pipeline avoids expansions that dilute reasoning, especially in large, dense KGs (e.g., SemMedDB).
\item \textbf{Chain-of-Thought Alignment:} Dividing queries into short, single-state steps (using "$\rightarrow$") and matching each with a subgraph query fosters accurate multi-hop solutions.
\item \textbf{Synergistic Components:} Removing CoT alignment or the final enhancement reduces accuracy, while fallback strategies ensure coverage if causal edges alone are insufficient.
\item \textbf{Cross-Model CoT Quality:} Pipeline success depends heavily on the coherence of the chain-of-thought. A weaker-model CoT can hamper a stronger model's inference, and vice versa.
\end{itemize}

Potential extensions include refining causal-mining algorithms, improving entity-linking to reduce retrieval misses, and integrating advanced domain modules. Additionally, applying these methods to legal or scientific literature QA could benefit from causal prioritization and CoT-driven retrieval. As knowledge graphs and LLMs evolve, the synergy between cause-effect edges and structured reasoning will become increasingly important.

\begin{credits}
\subsubsection{\ackname}
The contributing open-source communities are thanked for providing domain-specific resources such as SemMedDB, and the Hugging Face datasets for MedMCQA and MedQA. No external funding was received for this project.

\subsubsection{\discintname}
The authors declare no competing interests relevant to the content of this article.
\end{credits}

\bibliographystyle{splncs04}
\bibliography{reference}

\end{document}